\documentclass{article}

\usepackage[final]{corl_2018}

\usepackage{amssymb,amsmath,array}
\usepackage{url,balance}
\usepackage{times}
\usepackage{graphicx}
\usepackage{tabularx}
\usepackage{booktabs}
\usepackage{xcolor}
\usepackage{changebar}
\usepackage{subcaption}
\usepackage{natbib}
\title{Human Action Recognition and Assessment via \\Deep Neural Network Self-Organization}

\author{German I. Parisi\\
  Dept. of Computer Science, Universit\"at Hamburg, Germany\\
  \url{parisi@informatik.uni-hamburg.de}\\
  \url{http://giparisi.github.io}\\
}

\begin{document}
\maketitle

%===============================================================================

\begin{abstract}
The robust recognition and assessment of human actions are crucial in human-robot interaction (HRI) domains. While state-of-the-art models of action perception show remarkable results in large-scale action datasets, they mostly lack the flexibility, robustness, and scalability needed to operate in natural HRI scenarios which require the continuous acquisition of sensory information as well as the classification or assessment of human body patterns in real time.
In this chapter, I introduce a set of hierarchical models for the learning and recognition of actions from depth maps and RGB images through the use of neural network self-organization.
A particularity of these models is the use of growing self-organizing networks that quickly adapt to non-stationary distributions and implement dedicated mechanisms for continual learning from temporally correlated input.
\end{abstract}


%===============================================================================

\section{Introduction}
\label{chGPsec:1}

%Introduction to human action recognition and applications in HRI domains
%------------

\begin{figure*}[t]
\begin{centering}
\includegraphics[width=0.57\textwidth]{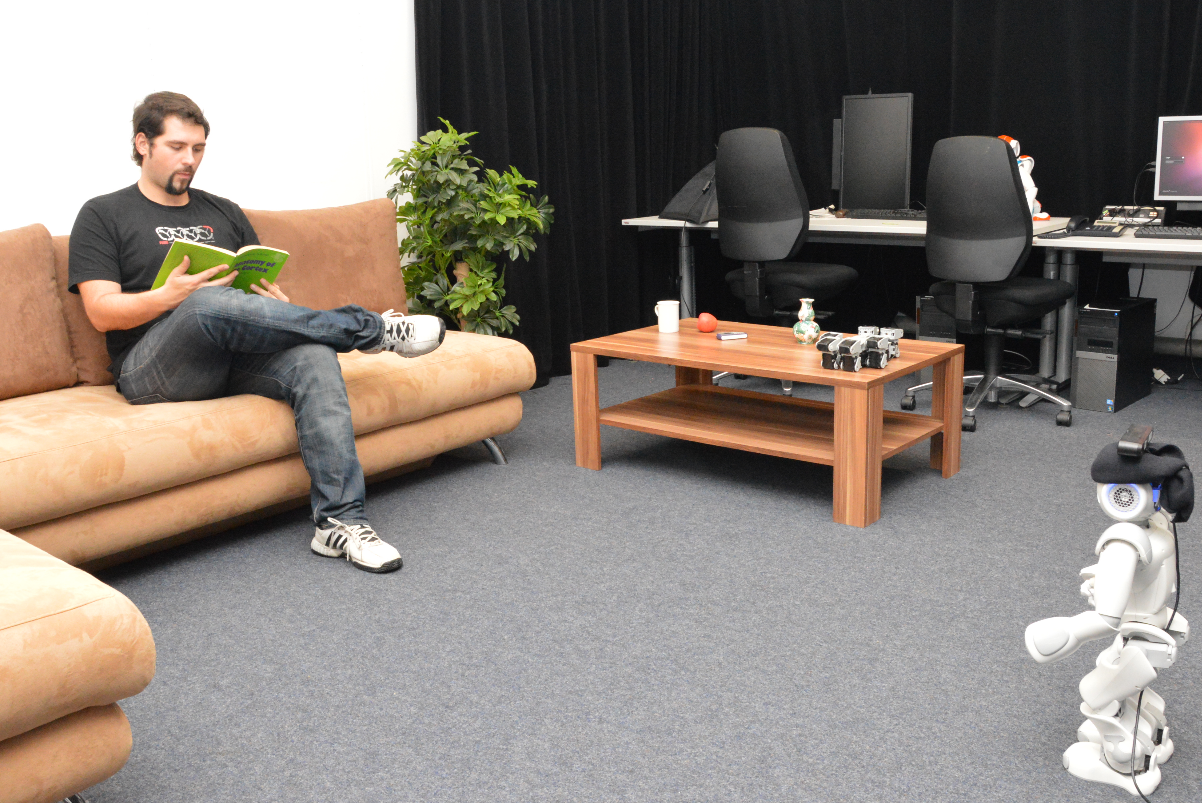}
\par\end{centering}
\caption[Person monitoring in a home-like environment]{\label{fig:fff}Person tracking and action recognition with a depth sensor on a humanoid robot in a domestic environment.~\citep{ParisiBook}.}
\end{figure*}

Artificial systems for human action recognition from videos have been extensively studied in the literature, with a large variety of machine learning models and benchmark datasets~\citep{Poppe2010,Guo2016}.
The robust learning and recognition of human actions are crucial in human-robot interaction (HRI) scenarios where, for instance, robots are required to efficiently process rich streams of visual input with the goal of undertaking assistive actions in a residential context (Fig.~\ref{fig:fff}).

Deep learning architectures such as convolutional neural networks (CNNs) have been shown to recognize actions from videos with high accuracy through the use of hierarchies that functionally resemble the organization of earlier areas of the visual cortex (see~\cite{Guo2016} for a survey).
However, the majority of these models are computationally expensive to train and lack the flexibility and robustness to operate in the above-described HRI scenarios.
A popular stream of vision research has focused on the use of depth sensing devices such as the Microsoft Kinect and ASUS Xtion Live for human action recognition in HRI applications using depth information instead of, or in combination with, RGB images.
Post-processed depth map sequences provide real-time estimations of 3D human motion in cluttered environments with increased robustness to varying illumination conditions and reducing the computational cost for motion segmentation and pose estimation~(see \cite{Han2013} for a survey).
However, learning models using low-dimensional 3D information (e.g. 3D skeleton joints) have often failed to show robust performance in real-world environments since this type of input can be particularly noisy and susceptible to self-occlusion.

In this chapter, I introduce a set of neural network models for the efficient learning and classification of human actions from depth information and RGB images.
These models use different variants of growing self-organizing networks for the learning of action sequences and real-time inference.
In Section~\ref{chGPsec:SO}, I summarize the fundamentals of neural network self-organization with focus on a particular type of growing network, the Grow When Required (GWR) model, that can grow and remove neurons in response to a time-varying input distribution, and the Gamma-GWR which extends the GWR with temporal context for the efficient learning of visual representations from temporally correlated input.
Hierarchical arrangements of such networks, which I describe in Section~\ref{chGPsec:AR}, can be used for efficiently processing body pose and motion features and learning a set of training actions.

%Use of SO networks for unsupervised learning?
%------------

%Emotion recognition
%-------------------

Understanding people's emotions plays a central role in human social interaction and behavior~\citep{Picard1997}.
Perception systems making use of affective information can significantly improve the overall HRI experience, for instance, by triggering pro-active robot behavior as a response to the user's emotional state.
An increasing corpus of research has been conducted in the recognition of affective states, e.g., through the processing of facial expressions~\citep{Fernando2013}, speech detection~\citep{Nwe2003} and the combination of these multimodal cues~\citep{Barros2016}.
While facial expressions can easily convey emotional states, it is often the case in HRI scenarios that a person is not facing the sensor or is standing far away from the camera, resulting in insufficient spatial resolution to extract facial features.
The recognition of emotions from body motion, instead, has received less attention in the literature but has a great value in HRI domains.
The main reason is that affective information is seen as harder to extrapolate from complex full-body expressions with respect to facial expressions and speech analysis.
In Section~\ref{chGPsec:ER}, I introduce a self-organizing neural architecture for emotion recognition from 3D body motion patterns.

%Body motion assessment
%----------------------

In addition to recognizing short-term behavior such as domestic daily actions and dynamic emotional states, it is of interest to learn the user's behavior over longer periods of time~\citep{Vettier2014}.
The collected data can be used to perform longer-term gait assessment as an important indicator for a variety of health problems, e.g., physical diseases and neurological disorders such as Parkinson's disease~\citep{Aerts2012}.
The analysis and assessment of body motion have recently attracted significant interest in the healthcare community with many application areas such as physical rehabilitation, diagnosis of pathologies, and assessment of sports performance.
The correctness of postural transitions is fundamental during the execution of well-defined physical exercises since inaccurate movements may not only significantly reduce the overall efficiency of the movement and but also increase the risk of injury~\citep{Kachouie2014}.
As an example, in the healthcare domain, the correct execution of physical rehabilitation routines is crucial for patients to improve their health condition~\citep{Velloso}.
Similarly, in weight-lifting training, correct postures improve the mechanical efficiency of the body and lead the athlete to achieve better results across training sessions.
In Section~\ref{chGPsec:MA}, I introduce a self-organizing neural architecture for learning body motion sequences comprising weight-lifting exercise and assessing their correctness in real time.

%Continual learning
%------------------

State-of-the-art models of action recognition have mostly proposed the learning of a static batch of body patterns~\citep{Guo2016}.
However, systems and robots operating in real-world settings are required to acquire and fine-tune internal representations and behavior in a continual learning fashion.
Continual learning refers to the ability of a system to seamlessly learn from continuous streams of information while preventing \textit{catastrophic forgetting}, i.e., a condition in which new incoming information strongly interferes with previously learned representations~\citep{Mermillod2013, Parisi2019}.
Continual machine learning research has mainly focused on the recognition of static image patterns whereas the processing of complex stimuli such as dynamic body motion patterns has been overlooked.
In particular, the majority of these models address supervised continual learning on static image datasets such as the MNIST~\citep{LeCun1998} and the CIFAR-10~\citep{Krizhevsky2009} and have not reported results on video sequences.
%In HRI scenarios, sequential input underlying spatiotemporal relations such as in the case of videos must be accounted for.
In Section~\ref{chGPsec:CL}, I introduce the use of deep neural network self-organization for the continual learning of human actions from RGB video sequences.
Reported results evidence that deep self-organization can mitigate catastrophic forgetting while showing competitive performance with state-of-the-art batch learning models.

Despite significant advances in artificial vision, learning models are still far from providing the flexibility, robustness, and scalability exhibited by biological systems.
In particular, current models of action recognition are designed for and evaluated on highly controlled experimental conditions, whereas systems and robots in HRI scenarios are exposed to continuous streams of (often noisy) sensory information.
In Section~\ref{chGPsec:OC}, I discuss a number of challenges and directions for future research.

\section{Neural Network Self-Organization}
\label{chGPsec:SO}

\subsection{Background}

Input-driven self-organization is a crucial component of cortical processing which shapes topographic maps based on visual experience~\citep{Willshaw1976,Nelson2000}.
Different artificial models of input-driven self-organization have been proposed to resemble the basic dynamics of Hebbian learning and structural plasticity~\citep{Hebb1949}, with neural map organization resulting from unsupervised statistical learning.
The goal of the self-organizing learning is to cause different parts of a network to respond similarly to certain input samples starting from an initially unorganized state.
Typically, during the training phase these networks build a map through a competitive process, also referred to as \textit{vector quantization}, so that a set of neurons represent prototype vectors encoding a submanifold in the input space.
Throughout this process, the network learns significant \textit{topological relations} of the input without supervision.

A well-established model is the self-organizing map (SOM)~\citep{Kohonen1991} in which the number of prototype vectors (or neurons) that can be trained is pre-defined.
However, empirically selecting a convenient number of neurons can be tedious, especially when dealing with non-stationary, temporally-correlated input distributions~\citep{Strickert2005}.
To alleviate this issue, a number of growing models have been proposed that dynamically allocate or remove neurons in response to sensory experience.
An example is the Grow When Required (GWR) network~\citep{Marsland2002} which grows or shrinks to better match the input distribution.
The GWR has the ability to add new neurons whenever the current input is not sufficiently matched by the existing neurons (whereas other popular models, e.g. Growing Neural Gas (GNG)~\citep{Fritzke1995}), will add neurons only at fixed, pre-defined intervals).
Because of their ability to allocate novel trainable resources, GWR-like models have the advantage of mitigating the disruptive interference of existing internal representations when learning from novel sensory observations.

\subsection{Grow When Required (GWR) Networks}

The GWR~\citep{Marsland2002} is a growing self-organizing network that learns the prototype neural weights from a multi-dimensional input distribution.
It consists of a set of neurons with their associated weight vectors and edges that create links between neurons.
For a given input vector $\textbf{x}(t)\in\mathbb{R}^n$, its best-matching neuron or unit (BMU) in the network, $b$, is computed as the index of the neural weight that minimizes the distance to the input:
\begin{equation}
b = \arg\min_{j\in A} \Vert \textbf{x}(t) - \textbf{w}_j \Vert,
\end{equation}
where $A$ is the set of neurons and $\Vert \cdot \Vert$ denotes the Euclidean distance.

The network starts with two randomly initialized neurons.
Each neuron $j$ is equipped with a habituation counter $h$ that considers the number of times that the neuron has fired.
Newly created neurons start with $h_j=1$ and iteratively decreased towards 0 according to the habituation rule
\begin{equation}\label{eq:FiringCounter}
\Delta h_i=\tau_i \cdot 1.05 \cdot (1-h_i)-\tau_i,
\end{equation}
where $i\in\{b,n\}$ and $\tau_i$ is a constant that controls the monotonically decreasing behavior.
Typically, $h_b$ is habituated faster than $h_n$ by setting $\tau_b>\tau_n$.

A new neuron is added if the activity of the network computed as $a=\exp{-\Vert \textbf{x}(t) - \textbf{w}_b \Vert}$ is smaller than a given activation threshold $a_T$ and if the habituation counter $h_b$ is smaller than a given threshold $h_T$.
The new neuron is created half-way between the BMU and the input.
This mechanism leads to creating neurons only after the existing ones have been sufficiently trained.

At each iteration, the neural weights are updated according to:
\begin{equation}
\Delta \textbf{w}_i = \epsilon_i \cdot h_i \cdot (\textbf{x}(t)-\textbf{w}_i),
\end{equation}
where $\epsilon_i$ is a constant learning rate ($\epsilon_n<\epsilon_b$) and the index $i$ indicates the BMU $b$ and its topological neighbors.
Connections between neurons are updated on the basis of neural co-activation, i.e. when two neurons fire together, a connection between them is created if it does not exist.

\begin{figure*}[t]
\centering
\includegraphics[width=0.9\textwidth]{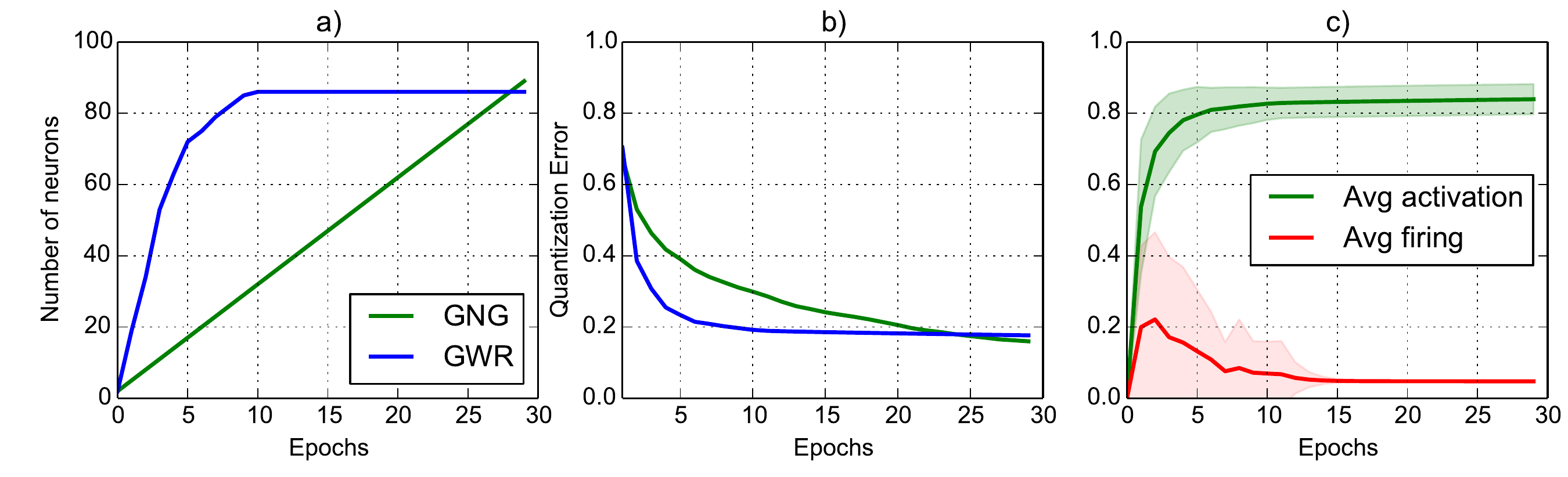}
\caption[Comparison of GNG and GWR]{Comparison of GNG and GWR training: a) number of neurons, b) quantization error, and c) GWR average activation and habituation counter through 30 training epochs on the Iris dataset~\citep{Parisi2017NN}.}% (\emph{$G^C$}).}
\label{fig:gnggwr}
\end{figure*}

While the mechanisms for creating new neurons and connections in the GWR do not resemble biologically plausible mechanisms of neurogenesis (e.g., \cite{Eriksson1998, Ming2011, Knoblauch2017}), the GWR learning algorithm represents an efficient model that incrementally adapts to non-stationary input.
%The GWR model creates new neurons whenever they are required and only after the training of existing ones.
A comparison between GNG and GWR learning in terms of the number of neurons, quantization error (average discrepancy between the input and its BMU), and parameters modulating network growth (average network activation and habituation rate) is shown in Fig.~\ref{fig:gnggwr}.
This learning behavior is particularly convenient for incremental learning scenarios since neurons will be created to promptly distribute in the input space, thereby allowing a faster convergence through iterative fine-tuning of the topological map.
The neural update rate decreases as the neurons become more habituated, which has the effect of preventing that noisy input interferes with consolidated neural representations.

\subsection{Gamma-GWR}

%Temporal Processing (GammaGWR)

The GWR model does not account for the learning of latent temporal structure.
For this purpose, the Gamma-GWR~\citep{Parisi2017NN} extends the GWR with temporal context.
Each neuron consists of a weight vector $\textbf{w}_j$ and a number $K$ of context descriptors $\textbf{c}_{j,k}$~($\textbf{w}_j,\textbf{c}_{j,k}\in\mathbb{R}^n$).

Given the input $\textbf{x}(t)\in\mathbb{R}^n$, the index of the BMU, $b$, is computed as:
\begin{equation} \label{eq:GetB}
b = \arg\min_{j\in A}(d_j),
\end{equation}
\begin{equation} \label{eq:BMUr}
d_j = \alpha_0 \Vert \textbf{x}(t) - \textbf{w}_j  \Vert + \sum_{k=1}^{K}\ \alpha_k \Vert \textbf{C}_k(t)-\textbf{c}_{j,k}\Vert,
\end{equation}
\begin{equation}\label{eq:MergeStep}
\textbf{C}_{k}(t) = \beta \cdot \textbf{w}_b^{t-1}+(1-\beta) \cdot \textbf{c}_{b,k-1}^{t-1},
\end{equation}
where $\Vert \cdot \Vert$ denotes the Euclidean distance, $\alpha_i$ and $\beta$ are constant values that modulate the influence of the temporal context, $\textbf{w}_b^{t-1}$ is the weight vector of the BMU at $t-1$, and $\textbf{C}_{k}\in\mathbb{R}^n$ is the global context of the network with $\textbf{C}_{k}(t_0)=0$.
If $K=0$, then Eq.~{\ref{eq:BMUr}} resembles the learning dynamics of the standard GWR without temporal context.
For a given input $\textbf{x}(t)$, the activity of the network, $a(t)$, is defined in relation to the distance between the input and its BMU (Eq.~\ref{eq:GetB}) as follows:
\begin{equation} \label{eq:Activity}
a(t)=\exp(-d_b),
\end{equation}
thus yielding the highest activation value of $1$ when the network can perfectly match the input sequence~($d_b=0$).

The training of the existing neurons is carried out by adapting the BMU $b$ and its neighboring neurons $n$:
\begin{equation}\label{eq:UpdateRateW}
\Delta \textbf{w}_i = \epsilon_i \cdot h_i \cdot (\textbf{x}(t) - \textbf{w}_i),
\end{equation}
\begin{equation}\label{eq:UpdateRateC}
\Delta \textbf{c}_{i, k} = \epsilon_i \cdot h_i \cdot (\textbf{C}_k(t) - \textbf{c}_{i, k}),
\end{equation}
where $i\in\{b,n\}$ and $\epsilon_i$ is a constant learning rate ($\epsilon_n<\epsilon_b$).
The habituation counters $h_i$ are updated according to Eq.~\ref{eq:FiringCounter}.

Empirical studies with large-scale datasets have shown that Gamma-GWR networks with additive neurogenesis show a better performance than a static network with the same number of neurons, thereby providing insights into the design of neural architectures in incremental learning scenarios when the total number of neurons is fixed~\citep{Parisi2018c}.

\section{Human Action Recognition}
\label{chGPsec:AR}

\subsection{Self-Organizing Integration of Pose-Motion Cues}
\label{chGPsec:SOI}

Human action perception in the brain is supported by a highly adaptive system with separate neural pathways for the distinct processing of body pose and motion features at multiple levels and their subsequent integration in higher areas~\citep{Ungerleider1982,Felleman1991}.
The ventral pathway recognizes sequences of body form snapshots, while the dorsal pathway recognizes optic-flow patterns.
Both pathways comprise hierarchies that extrapolate visual features with increasing complexity of representation~\citep{Taylor2015,Hasson2008,Lerner2011}.
It has been shown that while early visual areas such as the primary visual cortex (V1) and the motion-sensitive area (MT+) yield higher responses to instantaneous sensory input, high-level areas such as the superior temporal sulcus (STS) are more affected by information accumulated over longer timescales~\citep{Hasson2008}.
Neurons in higher levels of the hierarchy are also characterized by gradual invariance to the position and the scale of the stimulus~\citep{Orban1992}.
Hierarchical aggregation is a crucial organizational principle of cortical processing for dealing with perceptual and cognitive processes that unfold over time~\citep{Fonlupt2003}.
With the use of extended models of neural network self-organization, it is possible to obtain progressively generalized representations of sensory inputs and learn inherent spatiotemporal dependencies of input sequences.

\begin{figure*}[t]
\centering
\includegraphics[width=0.95\textwidth]{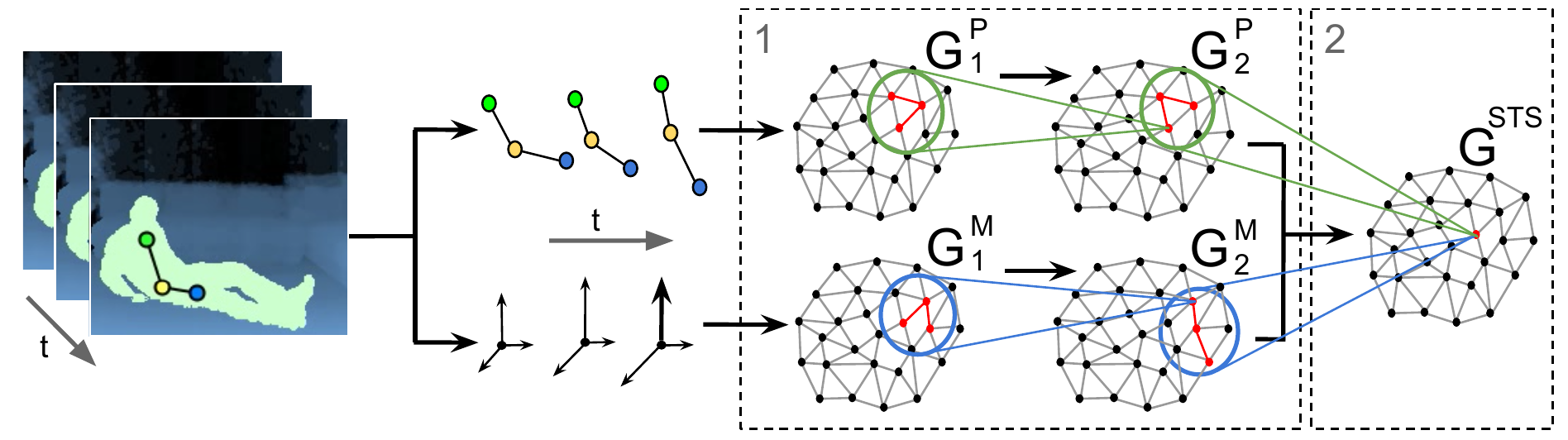}
\caption[GWR hierarchical architecture]{GWR-based architecture for pose-motion integration and action classification:
1) hierarchical processing of pose-motion features in parallel; % (\emph{$G^P$, $G^M$});
2) integration of neuron trajectories in the joint pose-motion feature space~\citep{parisi2015}.}% (\emph{$G^C$}).}
\label{fig:architectureGWR}
\end{figure*}

In \cite{parisi2015}, we proposed a learning architecture consisting of a two-stream hierarchy of GWR networks that processes extracted pose and motion features in parallel and subsequently integrates neuronal activation trajectories from both streams.
This integration network functionally resembles the response of STS model neurons encoding sequence-selective prototypes of action segments in the joint pose-motion domain.
An overall overview of the architecture is depicted in Fig.~\ref{fig:architectureGWR}.
The hierarchical arrangement of the networks yields progressively specialized neurons encoding latent spatiotemporal dynamics of the input.
We process the visual input under the assumption that action recognition is selective for temporal order~\citep{Poggio2003,Hasson2008}.
Therefore, the recognition of an action occurs only when neural trajectories are activated in the correct temporal order with respect to the learned action template.

Following the notation in Fig.~\ref{fig:architectureGWR}, $G^P_1$ and $G^M_1$ are trained with pose and motion features respectively.
After this step, we train $G^P_2$ and $G^M_2$ with concatenated trajectories of neural activations in the previous network layer.
The STS stage integrates pose-motion features by training G$^{STS}$ with the concatenation of vectors from $G^P_2$ and $G^M_2$ in the pose-motion feature space.
After the training of $G^{STS}$ is completed, each neuron will encode a sequence-selective prototype action segment, thereby integrating changes in the configuration of a person's body pose over time.
For the classification of actions, we extended the standard implementation of the GWR in which an associative matrix stores the frequency-based distribution of sample labels, i.e. each neuron stores the number of times that a given sample label has been associated to its neural weight.
This labeling strategy does not require a predefined number of action classes since the associative matrix can be dynamically expanded when a novel label class is encountered.

\begin{figure*}[t]
\centering
\includegraphics[width=0.75\textwidth]{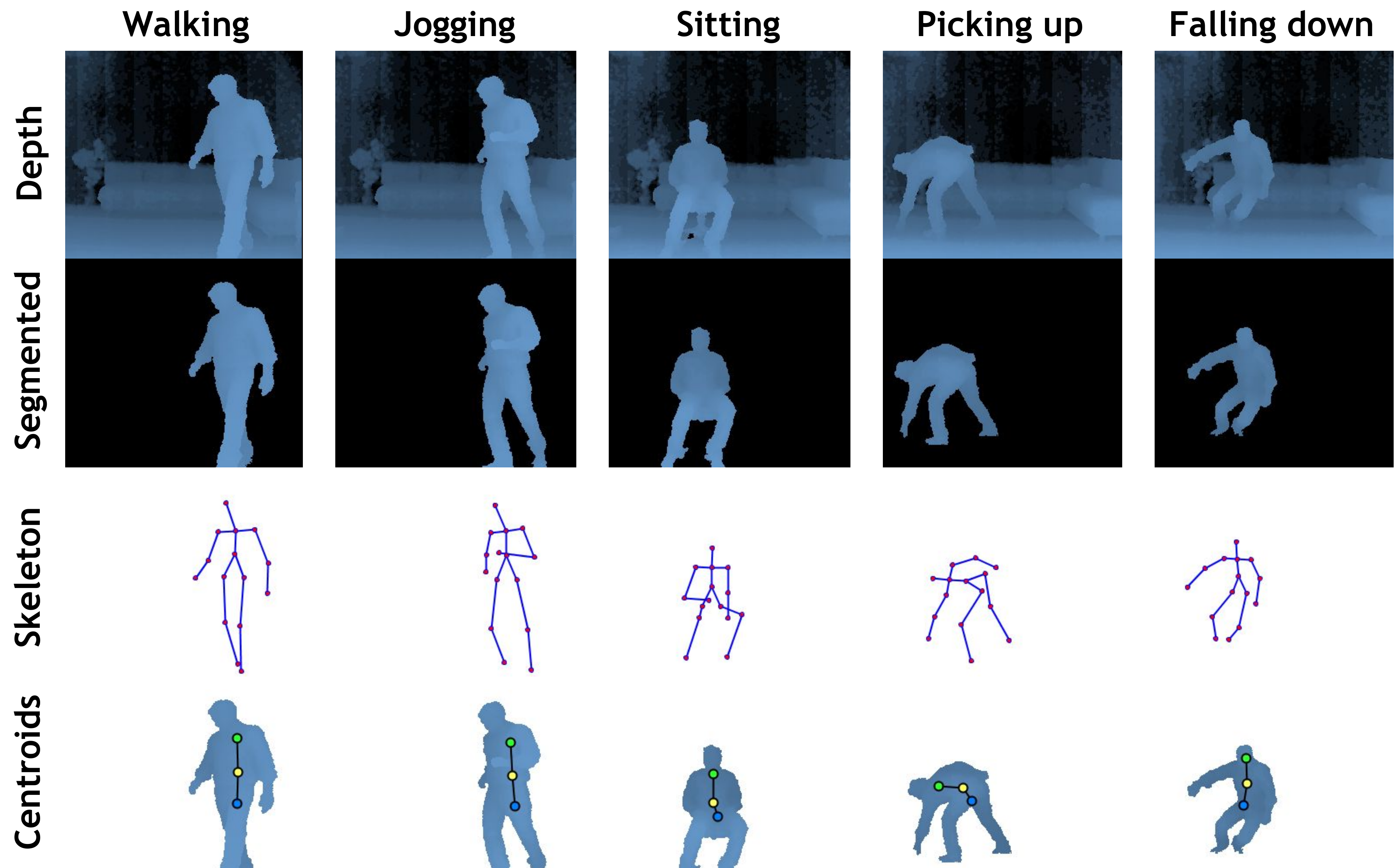}
\caption[Snapshots of actions from the KT dataset]{Snapshots of actions from the KT action dataset visualized as raw depth images, segmented body, skeleton, and body centroids.}% (\emph{$G^C$}).}
\label{chGPfig:actions}
\end{figure*}

\begin{figure*}[t!]
\centering
\includegraphics[width=0.59\textwidth]{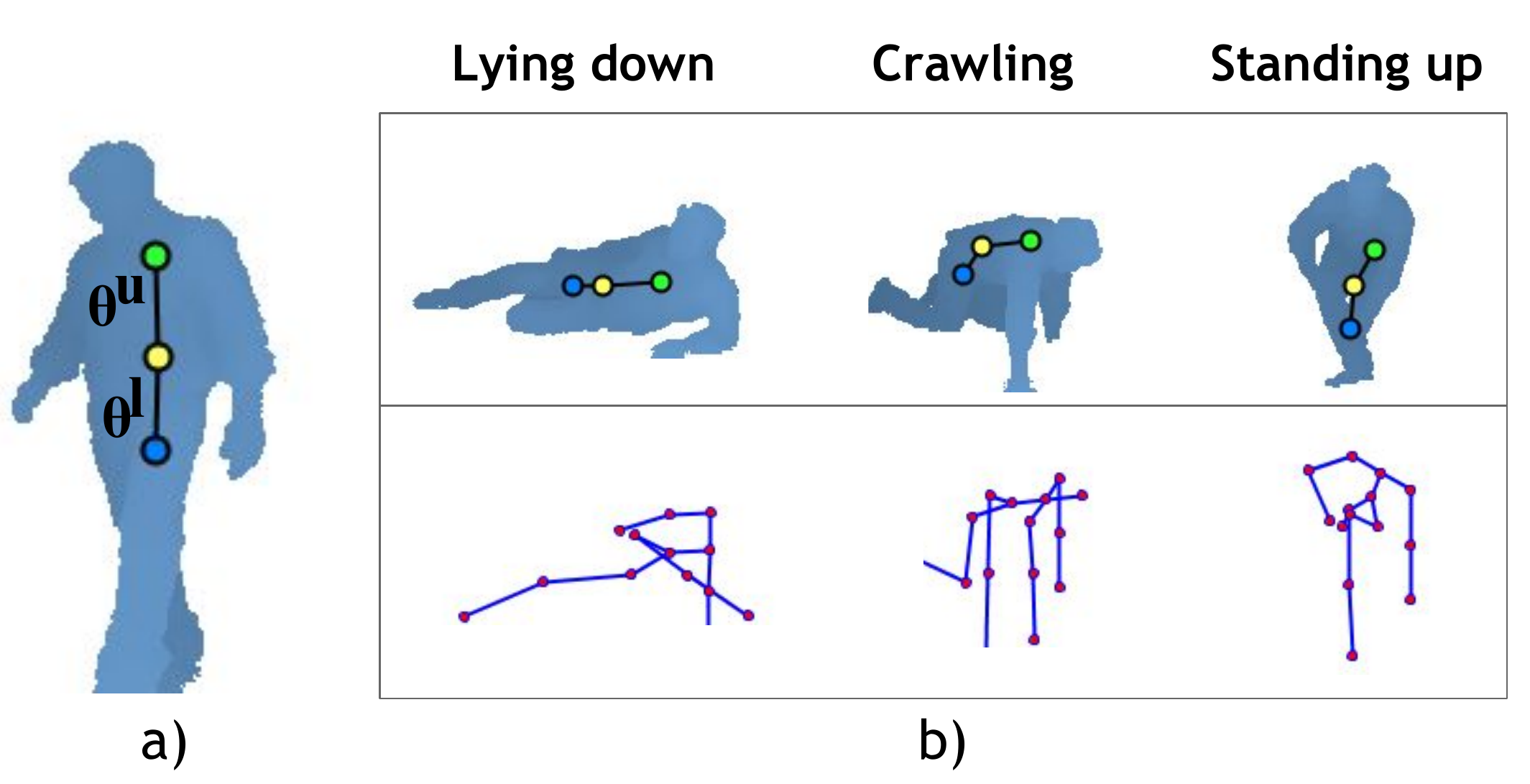}
\caption[Action representations]{Full-body action representations: (a) Three centroids with body slopes $\theta^u$ and $\theta^l$, and b) comparison of body centroids (top) and noisy skeletons (bottom).}% (\emph{$G^C$}).}
\label{fig:KTactionsNoise}
\end{figure*}

We evaluated our approach both on our Knowledge Technology (KT) full-body action dataset~\citep{Parisi2014} and the public action benchmark CAD-60~\citep{cad60_1}.
The KT dataset is composed of 10 full-body actions performed by 13 subjects with a normal physical condition.
The dataset contains the following actions: \textit{standing, walking, jogging, picking up, sitting, jumping, falling down, lying down, crawling,} and \textit{standing up}.
Videos were captured in a home-like environment with a Kinect sensor installed $1,30$ meters above the ground.
Depth maps were sampled with a VGA resolution of $640\times480$ and an operation range from $0.8$ to $3.5$ meters at $30$ frames per second.
From the raw depth map sequences, 3D body joints were estimated on the basis of the tracking skeleton model provided by OpenNI SDK.
Snapshots of full-body actions are shown in Fig.~\ref{chGPfig:actions} as raw depth images, segmented body silhouettes, skeletons, and body centroids.
We proposed a simplified skeleton model consisting of three centroids and two body slopes.
The centroids were estimated as the centers of mass that follow the distribution of the main body masses on each posture.
As can be seen in Fig.~\ref{fig:KTactionsNoise}, three centroids are sufficient to represent prominent posture characteristics while maintaining a low-dimensional feature space.
Such low-dimensional representation increases tracking robustness for situations of partial occlusion with respect to a skeleton model comprising a larger number of body joints.
Our experiments showed that a GWR-based approach outperforms the same type of architecture using GNG networks with an average accuracy rate of 94\% (5\% higher than GNG-based).

The Cornell activity dataset CAD-60 \citep{cad60_1} is composed of 60 RGB-D videos of four subjects (two males, two females, one left-handed) performing 12 activities: \textit{rinsing mouth, brushing teeth, wearing contact lens, talking on the phone, drinking water, opening pill container, cooking (chopping), cooking (stirring), talking on couch, relaxing on couch, writing on whiteboard, working on computer}.
The activities were performed in 5 different environments: office, kitchen, bedroom, bathroom, and living room.
The videos were collected with a Kinect sensor with distance ranges from 1.2 m to 3.5 m and a depth resolution of 640$\times$480 at 15 fps.
The dataset provides raw depth maps, RGB images, and skeleton data.
We used the set of 3D  positions without the \textit{feet}, leading to 13 joints (i.e., 39 input dimensions).
Instead of using world coordinates, we encoded the joint positions using the center of the hips as the frame of reference to obtain translation invariance.
We computed joint motion as the difference of two consecutive frames for each pose transition.

For our evaluation on the CAD-60, we adopted the same scheme as~\citep{cad60_1} using all the 12 activities plus a random action with a \textit{new person} strategy, i.e. the first 3 subjects for training and the remaining for test purposes.
We obtained 91.9\% precision, 90.2\% recall, and 91\% F-score.
The reported best state-of-the-art result is 93.8\% precision, 94.5\% recall, and 94.1\% F-score~\citep{cad60_6}, where they estimate, prior to learning, a number of key poses to compute spatiotemporal action templates.
Here, each action must be segmented into atomic action templates composed of a set of $n$ key poses, where $n$ depends on the action's duration and complexity.
Furthermore, experiments with real-time inference have not been reported.
The second-best approach achieves 93.2\% precision, 91.9\% recall, and 91.5\% F-score~\citep{cad60_5}, in which they used a dynamic Bayesian Mixture Model to classify motion relations between body poses.
However, the authors estimated their own skeleton model from raw depth images and did not use the one provided by the CAD-60 benchmark dataset.
Therefore, differences in the tracked skeleton exist that hinder a direct quantitative comparison with our approach.

\subsection{Emotion Recognition from Body Expressions}
\label{chGPsec:ER}

The recognition of emotions plays an important role in our daily life and is essential for social communication and it can be particularly useful in HRI scenarios.
For instance, a socially-assistive robot may be able to strengthen its relationship with the user if it can understand whether that person is bored, angry, or upset.
Body expressions convey an additional social cue to reinforce or complement facial expressions~\citep{Pollick2001}\citep{Sawada2003}.
Furthermore, this approach can complement the use of facial expressions when the user is not facing the sensor or is too distant from it for facial features to be computed.
Despite its promising applications in HRI domains, emotion recognition from body motion patterns has received significantly less attention with respect to facial expressions and speech analysis.

Movement kinematics such velocity and acceleration represent significant features when it comes to recognizing emotions from body patterns~\citep{Pollick2001}\citep{Sawada2003}.
Similarly, using temporal features in terms of body motion resulted in higher recognition rates than pose features alone~\citep{Patwardhan2016}.
Schindler \textit{et al.}~\citep{Schindler2008} presented an image-based classification system for recognizing emotion from images of body postures.
The overall recognition accuracy of his system resulted in 80\% for six basic emotions.
Although these systems show a high recognition rate, they are limited to postural emotions, which are not sufficient for a real-time interactive situation between humans and robots in a domestic environment.
Piana \textit{et al.}~\citep{Piana2014} proposed a real-time emotion recognition system using postural, kinematic, and geometrical features extracted from sequences of 3D skeletons videos.
However, they only considered a reduced set of upper-body joints, i.e., head, shoulders, elbows, hands, and torso.

\begin{figure*}[t]
\centering
\includegraphics[width=0.7\textwidth]{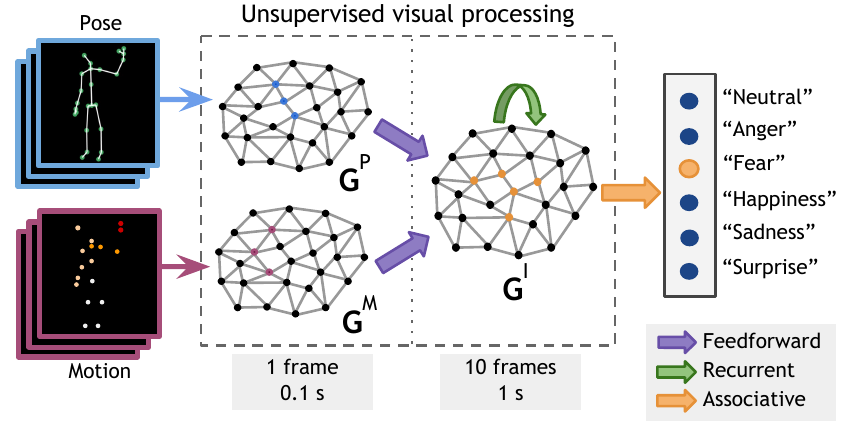}
{\caption{Proposed learning architecture with a hierarchy of self-organizing networks. The first layer processes pose and motion features from individual frames, whereas in the second layer a Gamma-GWR network learns the spatiotemporal structure of the joint pose-motion representations~\citep{Elfaramawy2017}.} \label{fig:ark}}% (\emph{$G^C$}).}
\end{figure*}

In \cite{Elfaramawy2017}, we proposed a self-organizing architecture to recognize emotional states from body motion patterns.
The focus of our study was to investigate whether full-body expressions from depth map videos convey adequate affective information for the task of emotion recognition.
The overall architecture, shown in Fig.~\ref{chGPfig:actions}, consists of a hierarchy of self-organizing networks for learning sequences of 3D body joint features.
In the first layer, two GWR networks~\citep{Marsland2002}, $G^P$ and $G^M$, learn a dictionary of prototype samples of pose and motion features respectively.
Motion features are obtained by computing the difference between two consecutive frames containing pose features.
In the second layer, a Gamma-GWR~\citep{Parisi2017NN}, $G^I$, is used to learn prototype sequences and associate symbolic labels to unsupervised visual representations of emotions for the purpose of classification.
While in the model presented in Section 3.1, networks were trained with concatenated trajectories of neural activations from a previous network layer, in this case we use the recurrent Gamma-GWR.
This is because sequences of bodily expressions comprising emotions require a larger temporal window to be processed and, by explicitly concatenating neural activations from previous layers, the dimensionality of the input increases~\citep{parisi2015}.
Here, instead, the temporal context of the Gamma-GWR is used to efficiently process larger temporal windows and reduce quantization error over time.
During the inference phase, unlabeled novel samples are processed by the hierarchical architecture, yielding patterns of neural weight activations.
One best-matching neuron in $G^I$ will activate for every 10 processed input frames.

For the evaluation of our system, we collected a dataset named the Body Expressions of Emotion (BEE), with nineteen participants performing six different emotional states: \textit{anger, fear, happiness, neutral, sadness}, and \textit{surprise}.
The dataset was acquired in an HRI scenario consisting of a humanoid robot Nao extended with a depth sensor to extract 3D body skeleton information in real time.
Nineteen participants took part in the data recordings (fourteen male, five female, age ranging from 21 to 33).
The participants were students at the University of Hamburg and they declared not to have suffered any physical injury resulting in motor impairments.
To compare the performance of our system to human observers, we performed an additional study in which 15 raters that did not take part in the data collection phase had to label depth map sequences as one of the six possible emotions.

For our approach, we used the full 3D skeleton model except for the \textit{feet}, leading to 13 joints (i.e., 39 input dimensions).
To obtain translation invariance, we encoded the joint positions using the center of the hips as the frame of reference.
We then computed joint motion as the difference of two consecutive frames for each pose transition.
Experimental results showed that our system successfully learned to classify the set of six training emotions and that its performance was very competitive with respect to human observers (see Table~\ref{tab:tablecc}).
The overall accuracy of emotions recognized by human observers was 90.2\%, whereas our system showed an overall accuracy of 88.8\%.

\begin{table}[t]
\begin{center}
    \begin{tabular}{ l*{8}{c}r}
      & \textbf{System} & \textbf{Human} \\ \hline
    Accuracy & 88.8\% & 90.2\% \\
    Precision & 66.3\% & 70.1\% \\
    Recall & 68\% & 70.7\% \\
    F-score &  66.8\% & 68.9\% \\ \hline
    \end{tabular}
    \caption{A comparison of overall recognition of emotions between our system and human performance.}~\label{tab:tablecc}
\end{center}
\end{table}

As additional future work, we could investigate the development of a multimodal emotion recognition scenario, i.e., by taking into account auditory information that complements the use of visual cues~\citep{Barros2016}.
The integration of audio-visual stimuli for emotion recognition has been shown to be very challenging but also strongly promising for a more natural HRI experience.

\section{Body Motion Assessment}
\label{chGPsec:MA}

\subsection{Background}

\begin{figure}[b]
\centering
\includegraphics[width=0.7\textwidth]{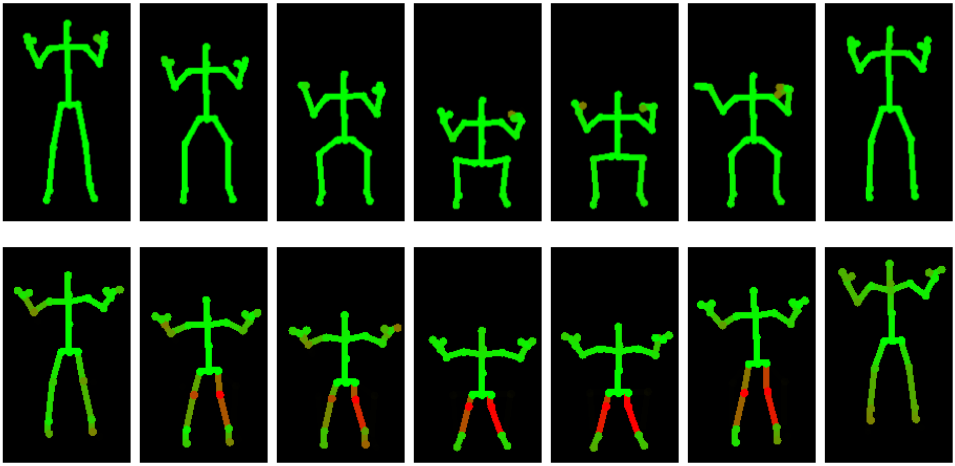}
\caption[Visual feedback for squat sequence]{Visual feedback for correct squat sequence (top), and a sequence
containing \textit{knees in} mistake (bottom; joints and limbs in red)~\citep{ParisiIJCNN2015}.}% (\emph{$G^C$}).}
\label{fig:kskeleton}
\end{figure}

The correct execution of well-defined movements plays a key role in physical rehabilitation and sports.
While the goal of action recognition approaches is to categorize a set of distinct classes by extrapolating inter-class differences, action assessment requires instead a model to capture intra-class dissimilarities that allow expressing a measurement on how much an action follows its learned template.
The quality of actions can be computed in terms of how much a performed movement matches the correct continuation of a learned motion sequence template.
Visual representations can then provide useful qualitative feedback to assist the user in the correct performance of the routine and the correction of mistakes~(Fig.~\ref{fig:kskeleton}).
The task of assessing the quality of actions and providing feedback in real time for correcting inaccurate movements represents a challenging visual task.

Artificial systems for the visual assessment of body motion have been previously investigated for applications mainly focused on physical rehabilitation and sports training.
For instance, Chan \textit{et al.}~\citep{Chan2011} proposed a physical rehabilitation system using a Kinect sensor for young patients with motor disabilities.
The idea was to assist the users while performing a set of simple movements necessary to improve their motor proficiency during the rehabilitation period.
Although experimental results have shown improved motivation for users using visual hints, only movements involving the arms at constant speed were considered.
Furthermore, the system does not provide real-time feedback to enable the user to timely spot and correct mistakes.
Similarly, Su \textit{et al.}~\citep{Su2013} proposed the estimation of feedback for Kinect-based rehabilitation exercises by comparing tracked motion with a pre-recorded execution by the same person.
The comparison was carried out on sequences using dynamic time warping and fuzzy logic with the Euclidean distance as a similarity measure.
The evaluation of the exercises was based on the degree of similarity between the current sequence and a correct sequence.
The system provided qualitative feedback on the similarity of body joints and execution speed, but it did not suggest the user how to correct the movement.

\subsection{Motion Prediction and Correction}

\begin{figure}[t]
\centering
\includegraphics[width=0.75\textwidth]{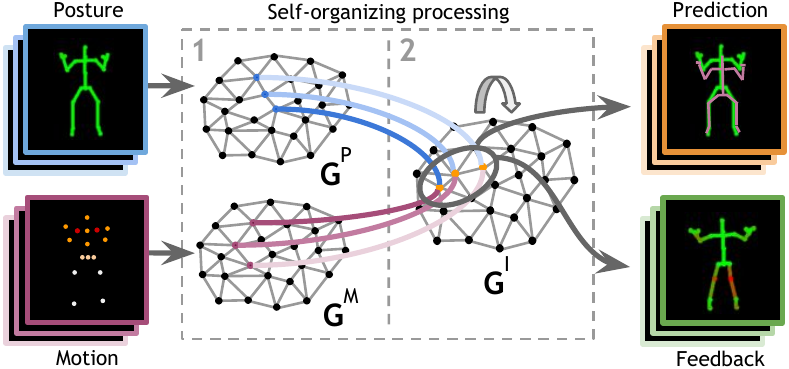}
\caption[Learning architecture for motion assessment]{Learning architecture with growing self-organizing networks. In Layer 1, two GWR networks learn posture and motion features respectively. In Layer 2, a Gamma-GWR learns spatiotemporal dynamics of body motion. This mechanism allows predicting the template continuation of a learned sequence and computing feedback as the difference between its current and its expected execution~\citep{ParisiROMAN2016}. }% (\emph{$G^C$}).}
\label{fig:architectureMergeGWR}
\end{figure}

In \cite{ParisiROMAN2016}, we proposed a learning architecture that consists of two hierarchically arranged layers with self-organizing networks for human motion assessment in real time~(Fig.~\ref{fig:architectureMergeGWR}).
The first layer is composed of two GWR networks, $G^P$ and $G^M$, that learn a dictionary of posture and motion feature vectors respectively.
This hierarchical scheme has the advantage of using a fixed set of learned features to compose more complex patterns in the second layer, where the Gamma-GWR $G^I$ with $K=1$ is trained with sequences of posture-motion activation patterns from the first layer to learn the spatiotemporal structure of the input.

The underlying idea for assessing the quality of a sequence is to measure how much the current input sequence differs from a learned sequence template.
Provided that a trained model $G^I$ represents a training sequence with a satisfactory degree of accuracy, it is then possible to quantitatively compute how much a novel sequence differs from such expected pattern.
We defined a function $\mathfrak{f}_\Omega$ that computes the difference of a current input sequence, $\Omega_t$, from its expected input, i.e. the prediction of the next element of the sequence given $\Omega_{t-1}$:
\begin{equation} \label{eq:feedback}
\mathfrak{f}_\Omega(t) = \Vert \Omega_t- \mathfrak{p}(\Omega_{t-1}) \Vert,
\end{equation}

\begin{equation} \label{eq:feedback2}
\mathfrak{p}(\Omega_{t-1}) = \textbf{w}_p ~\text{with} ~p={\arg\min_{j\in A} \Vert \textbf{c}_j - \Omega_{t-1} \Vert},
\end{equation}
where $A$ is the set of neurons and $\Vert \cdot \Vert$ denotes the Euclidean distance.
Since the weight and context vectors of the prototype neurons lie in the same feature space as the input ($\textbf{w}_i,\textbf{c}_i \in \mathbb{R}^{\left\vert{\Omega}\right\vert}$), it is possible to provide joint-wise feedback computations.
The recursive prediction function $\mathfrak{p}$ can be applied an arbitrary number of timesteps into the future.
Therefore, after the training phase is completed, it is possible to compute $\mathfrak{f}_\Omega(t)$ in real time with linear computational complexity $\mathcal{O}(\left\vert{A}\right\vert)$.

\begin{figure}[t]
\centering
\includegraphics[width=.7\textwidth]{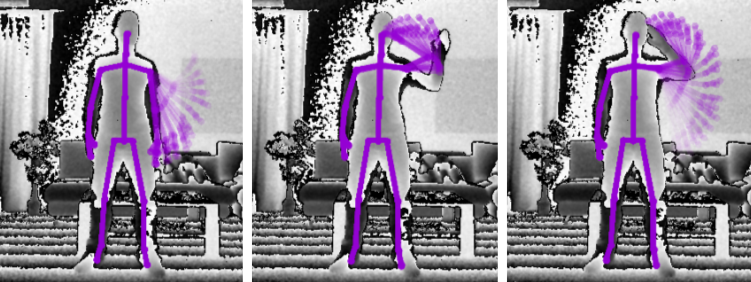}
\caption[Movement prediction for action assessment]{Visual hints for the correct execution of the \textit{Finger to nose} routine. Progressively fading violet lines indicate the learned action template~\citep{ParisiIJCNN2015}.}% (\emph{$G^C$}).}
\label{fig:nose}
\end{figure}

The visual effect of this prediction mechanism is shown in Fig.~\ref{fig:nose}.
For this example, the architecture was trained with the \textit{Finger to nose} routine which consists of keeping your arm bent at the elbow and then touching your nose with the tip of your finger.
As soon the person starts performing the routine, we can see progressively fading violet lines representing the next $30$ time steps which lead to visual assistance for successful execution.
The value $30$ was empirically determined to provide a substantial reference to future steps while limiting visual clutter.
To compute visual feedback, we used the $\mathfrak{p}$ predictions as hints on how to perform a routine over 100 timesteps into the future, and then use $\mathfrak{f}_\Omega(t)$ to spot mistakes on novel sequences that do not follow the expected pattern for individual joint pairs.
Execution mistakes are detected if $\mathfrak{f}_\Omega(t)$ exceeds a given threshold $\mathfrak{f}_T$ over $i$ timesteps.
Visual representations of these computations can then provide useful qualitative feedback to correct mistakes during the performance of the routine~(Fig.~\ref{fig:kskeleton}).
Our approach learns also motion intensity to better detect temporal discrepancies.
Therefore, it is possible to provide accurate feedback on posture transitions and the correct execution of lockouts.

\subsection{Dataset and Evaluation}

We evaluate our approach with a data set containing 3 powerlifting exercises performed by 17 athletes: High bar back squat, Deadlift, and Dumbbell lateral raise.
The data collection took place at the Kinesiology Institute of the University of Hamburg, Germany, where 17 volunteering participants (9 male, 8 female) performed 3 different powerlifting exercises.
We captured body motion of correct and incorrect executions with a Kinect v2 sensor and estimated body joints using Kinect SDK 2.0 that provides a set of $25$ joint coordinates at $30$ frames per second.
The participants executed the routines frontal to the sensor placed at 1 meter from the ground.
We extracted the 3D joints for \textit{head, neck, wrists, elbows, shoulders, spine, hips, knees}, and \textit{ankles}, for a total of 13 3D-joints (39 dimensions).
We computed motion intensity from posture sequences as the difference between consecutive joint pairs.
The Kinect's skeleton model (Fig.~\ref{fig:kskeleton}), although not faithful to human anatomy, provides reliable estimations of the joints' position over time when the user is facing the sensor.
We manually segmented single repetitions for all exercises.
In order to obtain translation invariance, we subtracted the $\textit{spine\_base}$ joint (the center of the hips) from all the joints in absolute coordinates.

We evaluated our method for computing feedback with individual and multiple subjects.
We divided the correct body motion data with 3-fold cross-validation into training and test sets and trained the models with data containing correct motion sequences only.
For the inference phase, both the correct and incorrect movements were used with feedback threshold $\mathfrak{f}^T=0.7$ over $100$ frames.
Our expectation was that the output of the feedback function would be higher for sequences containing mistakes.
We observed true positives (TP), false negatives (FN), true negatives (TN), and false positives (FP) as well as the measures true positive rate (TPR or sensitivity), true negative rate (TPR or specificity), and positive predictive value (PPV or precision).
Results for single- and multiple-subject data on E1, E2, and E3 routines are displayed in Table 6.1 and 6.2 respectively, along with a comparison with the best-performing feedback function $\mathfrak{f}_b$ from \citep{ParisiIJCNN2015} in which we used only pose frames without explicit motion information.

The evaluation on single subjects showed that the system successfully provides feedback on posture errors with high accuracy.
GWR-like networks allow reducing the temporal quantization error over longer timesteps, so that more accurate feedback can be computed and thus reduce the number of false negatives and false positives.
Furthermore, since the networks can create new neurons according to the distribution of the input, each network can learn a larger number of possible executions of the same routine, thus being more suitable for training sessions with multiple subjects.
Tests with multiple-subject data showed significantly decreased performance, mostly due to a large number of false positives.
This is not exactly a flaw due to the learning mechanism but rather a consequence people having different body configurations and, therefore, slightly different ways to perform the same routine.
To attenuate this issue, we can set different values for the feedback threshold $\mathfrak{f}_T$.
For larger values, the system would tolerate more variance in the performance.
However, one must consider whether a higher degree of variance is not desirable in some application domains.
For instance, rehabilitation routines may be tailored to a specific subject based on their specific body configuration and health condition.

\begin{table}[t]
\caption{Single-subject evaluation.}
\begin{center}
    \begin{tabular}{ l*{8}{c}r}
      &   & TP & FN & TN & FP & TPR & TNR & PPV \\ \hline
    E1 & $\mathfrak{f}_\mathfrak{b}$ & 35 & 10 & 33 & 0 & 0.77 & 1 & 1 \\
       & $\mathfrak{f}_\Omega$ & 35 & 2 & 41 & 0 & 0.97 & 1 & 1 \\ \hline
    E2 & $\mathfrak{f}_\mathfrak{b}$ & 24 & 0 & 20 & 0 & 1 & 1 & 1 \\
       & $\mathfrak{f}_\Omega$ & 24 & 0 & 20 & 0 & 1 & 1 & 1 \\ \hline
   E3 & $\mathfrak{f}_\mathfrak{b}$ & 63 & 0 & 26 & 0 & 1 & 1 & 1 \\
       & $\mathfrak{f}_\Omega$ & 63 & 0 & 26 & 0 & 1 & 1 & 1 \\ \hline
    \end{tabular}
\end{center}
\end{table}

\begin{table}[t]
\caption{Multi-subject evaluation. Best results in bold.}
\begin{center}
    \begin{tabular}{ l*{8}{c}r}
      &   & TP & FN & TN & FP & TPR & TNR & PPV \\ \hline
    E1 & $\mathfrak{f}_\mathfrak{b}$ & 326 & 1 & 7 & 151 & 0.99 & 0.04 & 0.68 \\
       & $\mathfrak{f}_\Omega$ & 328 & 1 & 13 & 143 & 0.99 & \textbf{0.08} & \textbf{0.70} \\ \hline
    E2 & $\mathfrak{f}_\mathfrak{b}$ & 127 & 2 & 0 & 121 & 0.98 & 0 & 0.51 \\
       & $\mathfrak{f}_\Omega$ & 139 & 0 & 0 & 111 & \textbf{1}    & 0 & \textbf{0.56} \\ \hline
   E3 & $\mathfrak{f}_\mathfrak{b}$ & 123 & 0 & 8 & 41 & 1     & 0.16 & 0.75 \\
       & $\mathfrak{f}_\Omega$ & 126 & 0 & 15 & 31 & 1     & \textbf{0.33} & \textbf{0.80} \\ \hline
    \end{tabular}
\end{center}
\end{table}

Our results encourage further work in embedding this type of real-time system into an assistive robot that can interact with the user and motivate the correct performance of physical rehabilitation routines and sports training.
The positive effects of having a motivational robot for health-related tasks has been shown in a number of studies~\citep{Dautenhahn1999,Kidd2007,Nalin2012}.
The assessment of body motion plays a role not only for the detection of mistakes on training sequences but also in the timely recognition of gait deterioration, e.g., linked to age-related cognitive declines.
Growing learning architectures are particularly suitable for this task since they can adapt to the user through longer periods of time while still detecting significant changes in their motor skills.

\section{Continual Learning of Human Actions}
\label{chGPsec:CL}

\subsection{Background}

%What is CL?
%-----------

%NN introduction
%---------------

Deep learning models for visual tasks typically comprise a set of convolution and pooling layers trained in a hierarchical fashion for yielding action feature representations with increasing degree of abstraction~(see \citep{Guo2016} for a recent survey).
This processing scheme is in agreement with neurophysiological studies supporting the presence of functional hierarchies with increasingly large spatial and temporal receptive fields along cortical pathways~\citep{Poggio2003, Hasson2008}
However, the training of deep learning models for action sequences has been proven to be computationally expensive and requires an adequately large number of training samples for the successful learning of spatiotemporal filters.
Consequently, the question arises whether traditional deep learning models for action recognition can account for real-world learning scenarios, in which the number of training samples may not be sufficiently high and system may be required to learn from novel input in a continual learning fashion.

Continual learning refers to the ability of a system to continually acquire and fine-tune knowledge and skills over time while preventing \textit{catastrophic forgetting}~(see \citep{Chen18, Parisi2019} for recent reviews).
Empirical evidence shows that connectionists architectures are in general prone to catastrophic forgetting, i.e., when learning a new class or task, the overall performance on previously learned classes and tasks may abruptly decrease due to the novel input interfering with or completely overwriting existing representations~\citep{French1999, Mermillod13}.
To alleviate catastrophic forgetting in neural networks, researchers have studied how to address the \textit{plasticity-stability dilemma}~\citep{Grossberg1980}, i.e. how which extent networks should adapt to novel knowledge without forgetting previously learned knowledge.
Specifically for self-organizing networks such as the GWR, catastrophic forgetting is modulated by the conditions of map plasticity, the available resources to represent information, and the similarity between new and old knowledge~\citep{Parisi2017NN, Richardson08}.
While the vast majority of the proposed continual learning models are designed for processing i.d.d. data from datasets of static images such as MNIST and CIFAR~(e.g.~\cite{Kirkpatrick17, Rebuffi16, Shin17, Zenke17}), here I introduce deep self-organization for the continual learning of non-stationary, non-i.d.d. data from videos comprising human actions. 

The approaches described in Section~\ref{chGPsec:AR} and ~\ref{chGPsec:MA} rely on the extraction of a simplified 3D skeleton model from which low-dimensional pose and motion features can be computed to process actor-independent action dynamics.
The use of such models is in line with biological evidence demonstrating that human observers are very proficient in learning and recognizing complex motion underlying a skeleton structure~\citep{Jastorff2006,Hiris2007}.
These studies show that the presence of a holistic structure improves the learning speed and accuracy of action patterns, also for non-biologically relevant motion such as artificial complex motion patterns.
However, skeleton models are susceptible to sensor noise and situations of partial occlusion and self-occlusion (e.g. caused by body rotation).
In this section, I describe how self-organizing architectures can be extended to learning and recognize actions in a continual learning fashion from raw RGB image questions.

\subsection{Deep Neural Network Self-Organization}

%NN approach
%-----------

\begin{figure*}[t]
\centering
\includegraphics[width=0.9\textwidth]{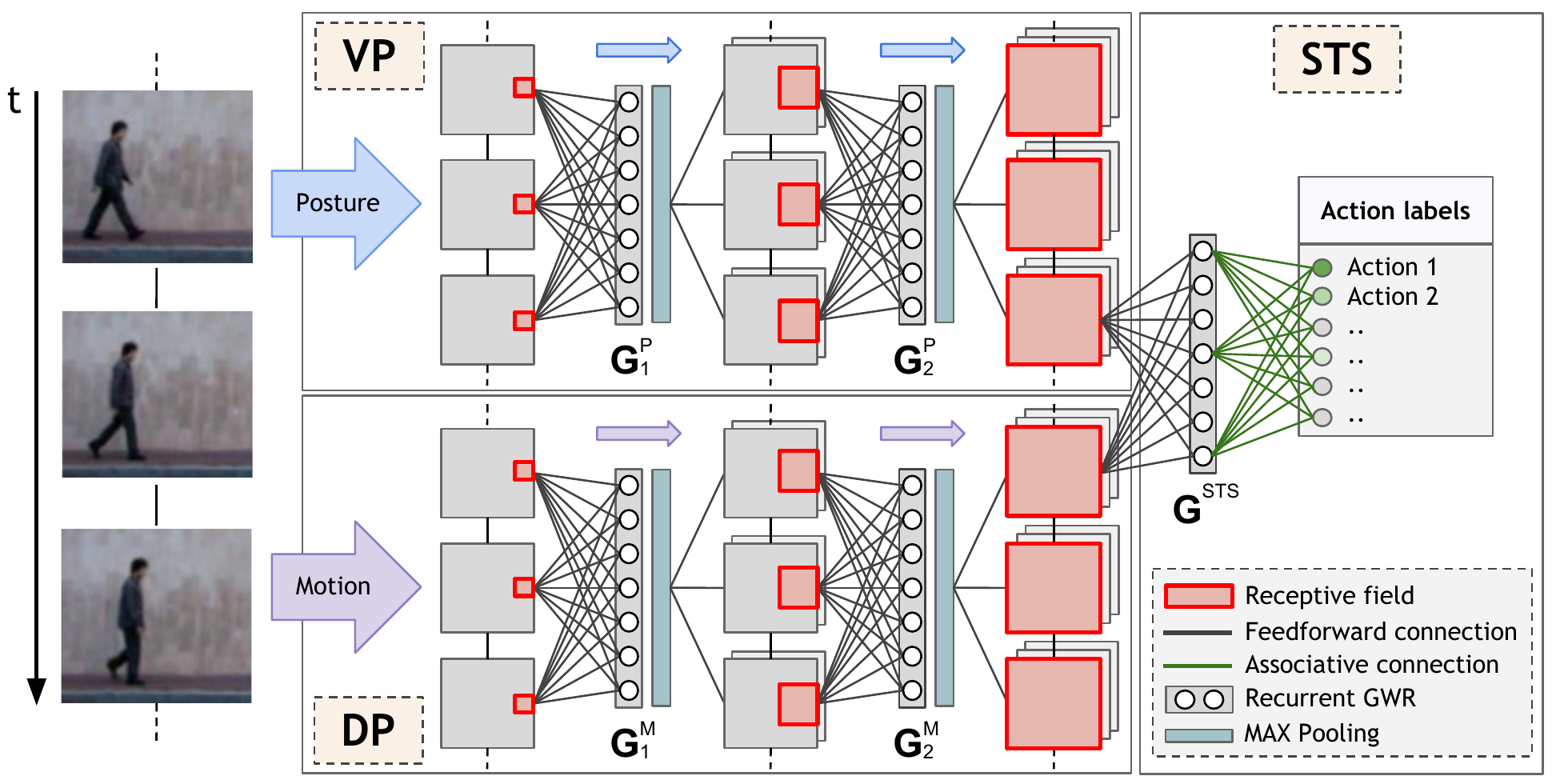}
\caption[Unsupervised deep learning architecture]{Diagram of our deep neural architecture with Gamma-GWR networks for continual action recognition. Posture and motion action cues are processed separately in the ventral (VP) and the dorsal pathway (DP) respectively. At the STS stage, the recurrent GWR network learns associative connections between prototype action representations and symbolic labels~\citep{Parisi2017NN}.}% (\emph{$G^C$}).}
\label{fig:dla}
\end{figure*}

In \cite{Parisi2017NN}, we proposed a self-organizing architecture consisting of a series of hierarchically arranged growing networks for the continual learning of actions from high-dimensional input streams (Fig.~\ref{fig:dla}).
Each layer in the hierarchy comprises a Gamma-GWR and a pooling mechanism for learning action features with increasingly large spatiotemporal receptive fields.
In the last layer, neural activation patterns from distinct pathways are integrated.
The proposed deep architecture is composed of two distinct processing streams for pose and motion features, and their subsequent integration in the STS layer.
Neurons in the G$^{STS}$ network are activated by the latest $K+1$ input samples, i.e. from time $t$ to $t-K$.

Deep architectures obtain invariant responses by alternating layers of feature detectors and nonlinear pooling neurons using, e.g., the maximum (MAX) operation, which has been shown to achieve higher feature specificity and more robust invariance with respect to linear summation~\citep{Guo2016}.
Robust invariance to translation has been obtained via MAX and average pooling, with the MAX operator showing faster convergence and improved generalization~\citep{Scherer2010}.
In our architecture, we implemented MAX-pooling layers after each Gamma-GWR network~(see Fig.~\ref{fig:dla}).
For each input image patch, a best-matching neuron $\textbf{w}_b^{(n-1)} \in \mathbb{R}^m$ is be computed in layer $n-1$ and only its maximum weight value $\widetilde{\textbf{w}}^{(n)}\in \mathbb{R}$ will be forwarded to the next layer $n$:
\begin{equation}\label{eq:pcapooling} 
\widetilde{\textbf{w}}^{(n)}=\max_{0 \leq i \leq m} \textbf{w}_{b,i}^{(n-1)},
\end{equation}
where $b$ is computed according to Eq.~\ref{eq:GetB} and the superscript on $\widetilde{\textbf{w}}^{(n)}$ indicates that this value is not an actual neural weight of layer $n$, but rather a pooled activation value from layer $n-1$ that will be used as input in layer $n$.
Since the spatial receptive field of neurons increases along the hierarchy, this pooling process will yield scale and position invariance.

\subsection{Datasets and Evaluation}

%NN results
%----------

We conducted experimental results with two action benchmarks: the Weizmann~\citep{Gorelick2005} and the KTH~\citep{Schuldt2004} datasets.

The Weizmann dataset contains 90 low-resolution image sequences with 10 actions performed by 9 subjects.
The actions are \textit{walk, run, jump, gallop sideways, bend, one-hand wave, two-hands wave, jump in place, jumping jack}, and \textit{skip}.
Sequences are sampled at $180\times144$ pixels with a static background and are about 3 seconds long.
We used aligned foreground body shapes by background subtraction included in the dataset. %(Fig.~\ref{fig:wwmas1}).
For compatibility with \citep{Schindler2008}, we trimmed all sequences to a total of 28 frames, which is the length of the shortest sequence, and evaluated our approach by performing \textit{leave-one-out} cross-validation, i.e., 8 subjects were used for training and the remaining one for testing.
This procedure was repeated for all 9 permutations and the results were averaged.
%Results for the recognition of 10-frame snippets are shown in Table~\ref{Tab:weit}.
Our overall accuracy was 98.7\%, which is competitive with the best reported result of 99.64\%~\citep{Gorelick2005}.
In their approach, they extracted action features over a number of frames by concatenating 2D body silhouettes in a space-time volume and used nearest neighbors and Euclidean distance to classify.
Notably, our results outperform the overall accuracy reported by \citep{jung2015} with three different deep learning models: convolutional neural network (CNN, 92.9\%), multiple spatiotemporal scales neural network (MSTNN, 95.3\%), and 3D CNN (96.2\%).
However, a direct comparison of the above-described methods with ours is hindered by the fact that they differ in the type of input and number of frames per sequence used during the training and the test phase.

The KTH action dataset contains 25 subjects performing 6 different actions: \textit{walking, jogging, running, boxing, hand-waving} and \textit{hand-clapping}, for a total of 2391 sequences.
Action sequences were performed in 4 different scenarios: indoor, outdoor, variations in scale, and changes in clothing.
Videos were collected with a spatial resolution of $160 \times 120$ pixels taken over homogeneous backgrounds and sampled at $25$ frames per second.
Following the evaluation schemes from the literature, we trained our model with 16 randomly selected subjects and used the other 9 subjects for testing.
The overall classification accuracy averaged across 5 trials achieved by our model was 98.7\%, which is competitive with the two best reported results: $95.6\%$~\citep{Ravanbakhsh15} and $95.04\%$~\citep{Gao2010}.
In the former approach, they used a hierarchical CNN model to capture sub-actions from complex ones.
%Higher levels of the hierarchy represent a coarse capture of an action sequence and lower levels represent fine action elements.
Key frames were extracted using binary coding of each frame in a video which helps to improve the performance of the hierarchical model (from 94.1\% to 95.6\%).
%To be noted is that the reported accuracy represents the result of the best trial, while an averaged accuracy across multiple trials has not been reported by the authors.
In the latter approach, they computed handcrafted interest points with substantial motion, which requires high computational requirements for the estimation of ad-hoc interest points.
%To be noted is that the results reported by~\citep{Gao2010} correspond to the average on the five best runs over a total of 30 trials, while the classification accuracy decreases to $90.93\%$ if averaging the five worst ones.
Our model outperforms other hierarchical models that do not rely on handcrafted features, such as 3D CNN ($90.2\%$, \citep{Ji2013}) and 3D CNN in combination with long short-term memory ($94.39\%$, \citep{Baccouche2011}).

\section{Conclusions and Open Challenges}
\label{chGPsec:OC}

%Summary of advantages and possible limitations
%-------

The underlying neural mechanisms for action perception have been extensively studied, comprising cortical hierarchies for processing body motion cues with increasing complexity of representation~\citep{Taylor2015,Hasson2008,Lerner2011}, i.e. higher-level areas process information accumulated over larger temporal windows with increasing invariance to the position and the scale of stimuli.
Consequently, the study of the biological mechanisms for action perception is fundamental for the development of artificial systems aimed to address the robust recognition of actions and learn in a continual fashion in HRI scenarios~\citep{ParisiRethink}.

Motivated by the process of input-driven self-organization exhibited by topographic maps in the cortex~\citep{Nelson2000,Willshaw1976,Miikkulainen2005}, I introduced learning architectures hierarchically arranged growing networks that integrate body posture and motion features for action recognition and assessment.
The proposed architectures can be considered a further step towards more flexible neural network models for learning robust visual representations on the basis of visual experience.
Successful applications of deep neural network self-organization include human action recognition~\citep{Parisi2014,parisi2015,Elfaramawy2017}, gesture recognition~\citep{ParisiFinger,ParisiHandSOM}, body motion assessment~\citep{ParisiIJCNN2015,ParisiROMAN2016}, human-object interaction~\citep{Mici2017,Mici2018}, continual learning~\citep{Parisi2017NN,Parisi2018b}, and audio-visual integration~\citep{Parisi2016J}.

Models of hierarchical action learning are typically feedforward.
However, neurophysiological studies have shown that the visual cortex exhibits significant feedback connectivity between different cortical areas~\citep{Felleman1991, Salin1995}.
In particular, action perception demonstrates strong top-down modulatory influences from attentional mechanisms~\citep{Thornton2002} and higher-level cognitive representations such as biomechanically plausible motion~\citep{Shiffrar1990}.
Spatial attention allows animals and humans to process relevant environmental stimuli while suppressing irrelevant information.
Therefore, attention as a modulator in action perception is also desirable from a computational perspective, thereby allowing the suppression of uninteresting parts of the visual scene and thus simplifying the detection and segmentation of human motion in cluttered environments.

The integration of multiple sensory modalities such as vision and audio is crucial for enhancing the perception of actions, especially in situations of uncertainty, with the aim to reliably operate in highly dynamic environments~\citep{Parisi2018iros}.
Experiments in HRI scenarios have shown that the integration of audio-visual cues significantly improves performance with respect to unimodal approaches for sensory-driven robot behavior~\citep{ParisiBook,Cruz2016,CruzIROS}.
The investigation of biological mechanisms of multimodal action perception is an important research direction for the development of learning systems exposed to rich streams of information in real-world scenarios.

\subsection*{Acknowledgements}
The author would like to thank Pablo Barros, Doreen Jirak, Jun Tani, and Stefan Wermter for great discussions and feedback.

\bibliographystyle{agsm}
\bibliography{chapter}  % .bib

\end{document}